\documentclass{svproc}
\usepackage{stfloats}
\usepackage{graphicx}
\usepackage{multirow}
\usepackage{amsmath}
\usepackage{caption}
\usepackage{float}
\usepackage{cuted}
\usepackage{subfigure}
\usepackage{xcolor}
\usepackage[linesnumbered,ruled,vlined]{algorithm2e}
\SetKwInput{KwInput}{Input}      
\SetKwInput{KwOutput}{Output}    
\usepackage{algorithmic}

\begin{document}

	\pagestyle{plain}
\mainmatter              
\title{Hybrid Machine Learning Models of Classifying Residential Requests for Smart Dispatching
}

\author{Tianen Chen \and Jincheng Sun \and Hongyi Lin \and Yan Liu \inst{*}}

\institute{Concordia University, Montreal QC H3G 1M8, Canada,\\
	\email{yan.liu@concordia.ca}}

\maketitle              

\begin{abstract}
This paper presents a hybrid machine learning method of classifying residential requests in natural language to responsible departments that provide timely responses back to residents under the vision of digital government services in smart cities. Residential requests in natural language descriptions cover almost every aspect of a city's daily operation. Hence the responsible departments are fine-grained to even the level of local communities. There are no specific general categories or labels for each request sample. This causes two issues for supervised classification solutions, namely 1) the request sample data is unbalanced and 2) lack of specific labels for training. To solve these issues, we investigate a hybrid machine learning method that generates meta-class labels by means of unsupervised clustering algorithms; applies two-word embedding methods with three classifiers (including two hierarchical classifiers and one residual convolutional neural network); and selects the best performing classifier as the classification result. We demonstrate our approach performing better classification tasks compared two benchmarking machine learning models, Naive Bayes classifier and a Multiple Layer Perceptron (MLP). In addition, the hierarchical classification method provides insights into the source of classification errors. 
\end{abstract}

\vspace{-0.3in}

\section{Introduction}
\label{intro}
The context of smart cities relates to the capability of analyzing and responding to specific requests from residents. In addition to social media networks such as Twitter and Chinese Weibo, an important source of city events down to the county level are service requests directly reported from metropolitan residents. These requests are through phone calls, emails and chatbots sent to metropolitan residential service centers that employ Human agents to dispatch the requests manually to corresponding service sectors for handling such issues. However, the processing capacity of call centers is bounded by the limitation of human operation. Therefore an artificial intelligence-enabled system helps to automatically convert audio-based reported requests into text content and then dispatches the request to the corresponding sector of services. Such automation improves service responsiveness.  

The core of such an automated request analysis and dispatching system is a machine learning model that classifies a request in natural language to an organization that is responsible for handling the case. To scope the research problem, this paper has the assumption that requests are all text-based. Any audio request is converted to the text-based content. The issues of classifying these text-based request data are two aspects. First, the labels for classification require processing on the raw datasets to produce suitable labels. It could be assumed that the field called the responsible department in the original dataset should be the labels. However, the responsible departments are also presented in natural language description that contains abbreviation, location information and less precise rename of a department. In addition, future requests may even result in new department names that are not part of the historical labels. Another issue is the request distribution over the corresponding responsible departments by nature are not evenly distributed. Some responsible departments are of small request cases and lead to the overall datasets unbalanced. 

In this paper, we propose a hybrid method to address the above two issues. Our method consists of (1) an NLP processing workflow for feature extraction; (2) a clustering algorithm for generating meta-classes for hierarchical training; (3) multiple classifiers with a Naive Bayes model, a fully connected neural network model and a residual neural network model to produce an optimal inference  for a certain request. Training the classification with generated meta-classes, we gain insights into the classification errors. 

A use case of our hybrid method has been developed on over 80,000 sample requests in Chinese that involve 157 responsible departments. Our method achieves with testing precision of 76.42\%  and log loss value of 1.192 over 35,663 test data that collected in different time period than the training datasets. Our code and samples of the dataset are open-sourced on Github\footnote{https://github.com/OneClickDeepLearning/classificationOfResidentialRequests}.

The \textbf{contribution} of this paper is three-fold as follows: 
\begin{itemize}
    \item We build an NLP-based feature engineering workflow with a rigorous measure of feature prediction power using information values; 
\item We demonstrate a hybrid machine learning method with both unsupervised and supervised machine learning to classify residential requests over unbalanced data sample distribution; 
\item We develop three classifiers to ensemble the best classification result. We compare the classification performance with two benchmarking models.
	
\end{itemize}

The structure of this paper is organized as follows: Section 2 presents the related work. In Section ~\ref{sub:featureextraction}, we introduce our dataset and feature engineering techniques. Section~\ref{sec:Hierarchical} describes a hierarchical classification method of generating meta-classes for classification of a large number of classes. The hybrid machine learning models in Section~\ref{sec:hybrid}. Finally, we present our assessment metrics and experiment results of the classification performance in Section~\ref{sec:evaluation}. We conclude our paper in Section~\ref{sec:conclusion}.

\section{Related Work}
\label{sec:1}

\textbf{Learning from few labeled data.} Kamal et al. proposed an Expectation-Maximization based method~\cite{nigam2000text} to train a classifier with few labeled data and use the classifier to label the high-confidence unlabelled data, then use the labeled data to train a new classifier, and repeat this process until the classifier converge. Blum et al.~\cite{blum1998combining} proposed a co-training method that allows learning from two views of the features. Rajat et al. proposed a Self-taught method that uses Auto-Encoder to learn higher-level representations with unlabelled data~\cite{raina2007self}.

\textbf{Imbalance dataset.} Nitesh et al. presented a method of SMOTE (Synthetic Minority Over-sampling Technique) \cite{chawla2002smote} to over-sampling the minority class by creating synthetic minority samples that fall between the minority data and their nearest neighbors. If the distance between the minority data and the neighbors is far, the synthetic data will have a large range of noise. Hui et al. presented a Borderline-SMOTE method \cite{han2005borderline} ensures that the sampling happens only when the majority of the selected neighbors are minority data to lower the randomness of noise.

\textbf{Ensemble of Classifiers.} Breiman proposed a bootstrap aggregating method\cite{breiman1996bagging} using bootstrap to extract several sub training sets from the original training set and train a classifier for each sub training set. Then each classifier gives a weighted vote for the classification. Yoav and Robert proposed a boost-based method called AdaBoost that uses weighted data to train weak classifiers. The data miss-classified by a weak classifier gains more weight to train the next weak classifier. The weak classifiers are weighted to form a confident classifier. Based on AdaBoost \cite{freund1997decision} method, Wei et al. proposed a cost-sensitive Boosting \cite{fan1999adacost} that increases the weight of misclassified training data that have a higher cost.

\textbf{Hierarchical Classification.} Pei-Yi et al proposed a hierarchically SVM text classification method that split a problem into sub-problems in the classification tree that can be solved accurately and efficiently \cite{hao2007hierarchically}. A hierarchical softmax architecture \cite{Bengio2015} was proposed by Morin and Bengio when dealing with a huge number of output classes. It significantly speeds up the training time compared to feed-forward networks.

\textbf{Convolutional Neural Networks on NLP Tasks.} Convolutional Neural Network (CNN) models have been demonstrated performing well in the NLP tasks of sentence classification~\cite{Cybenko1989}~\cite{Zhang:2015:CCN:2969239.2969312}. A CNN model consists of layers of convolutions with non-linear activation function such as \textit{ReLU} or \textit{tanh}.  In a CNN model, convolutions over the input layer are used to compute the output. As illustrated by Zhang~\cite{DBLP:journals/corr/ZhangW15b}, each region of the input is connected to a neuron in the output. Then each layer applies a filter to detect higher-level features. These features further go through a pooling layer to form a univariate feature vector for the penultimate layer. The final softmax layer then receives this feature vector as input and uses it to classify the sentence. 

Kim applied a single convolutional layer on top of Word2Vec embeddings \cite{kim2014convolutional} of words in a sentence to perform classification tasks. His work proves that convolutional neural network, with a good representation of the words, can provide promising results on NLP problems. Conneau et al. proposed ResNet-like deep convolutional neural network \cite{vdconvfortex} that can learn the different hierarchical representation of the input text. They use small convolutions and let the network learn what is the best combination of these extracted features. Their work allows the network to be deeper and finer-grained. Prabowo and Thelwall proposed a series of hybrid classifiers \cite{sentimentanalysis} that combine different rule-based classifiers and SVM as a pipeline to solve sentiment analysis problems and achieved good performance.

\section{Feature Engineering}\label{sub:featureextraction}
Feature engineering processes and transforms the dataset in texts to word vectors as inputs of machine learning models. The original dataset contains eight features with \textit{id}, \textit{time stamp}, four \textit{categories} of responsible departments, \textit{request description}, and \textit{responsible department description}. 

\begin{figure*}[h]
	\includegraphics[width=1.0\textwidth]{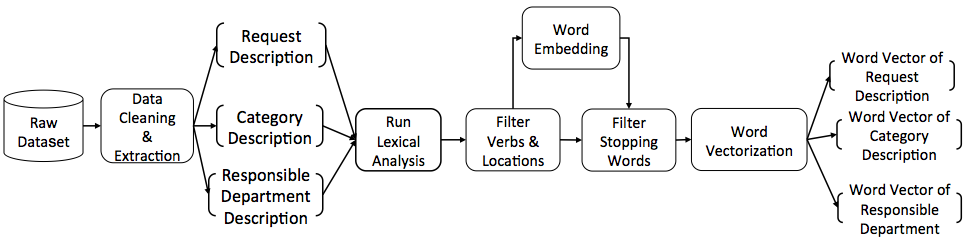}
	\caption{The major steps of feature engineering on two features of request description and responsible department description}
	\label{fig:featureeng-workflow}
\end{figure*}

The feature engineering workflow is depicted in Figure~\ref{fig:featureeng-workflow}. Feature engineering first removes invalid data samples in which the values of the responsible department description are missing or originally marked as non-available. The training dataset contains 849,861 records, including 145,542 records originally marked as invalid and 58,394 records without a responsible department description. Finally, valid dataset contains 645,924 records. 

\vspace{-0.3in}

\vspace{-0.4in}
\subsection{Data Preprocessing} 
Sentences are segmented into tokens before feature extraction. Two major methodologies of tokenization and segmentation are dictionary-based and statistics-based. The dictionary-based methods recognize words based on a maintained vocabulary  \cite{mikolov2013efficient}, while the statistics-based approach uses corpus as a resource to build a word-segmentation model. In this paper, we process the tokenization and segmentation with the second method using a tool called LTP, a Chinese language technology platform~\cite{LTP}. LTP consists of six Chinese processing modules, including 
1) Word Segmentation (WordSeg); 
2)Part-of-Speech Tagging (POSTag); 
3) Named Entity Recognition (NER); 
4) Word Sense Disambiguation (WSD); 
5) Syntactic Parsing (Parser) and Semantic Role Labeling (SRL). 

Further on, the segmented tokens are filtered by lexical analysis modules of LTP to eliminate tokens that include digits, words in other languages, punctuation, and stop words. A combined list of public available stopping-words is applied to the LTP tool. In addition, verbs, adjectives, adverbs are also excluded. Organization names and location-relevant nouns consist of more than one tokens. The NER module of LTP recognizes and merges these nouns into single words. 

\subsection{Data Distribution} \label{sub:datadistribution}
\begin{figure}[h]
	\centering
	\includegraphics[width=0.7\textwidth]{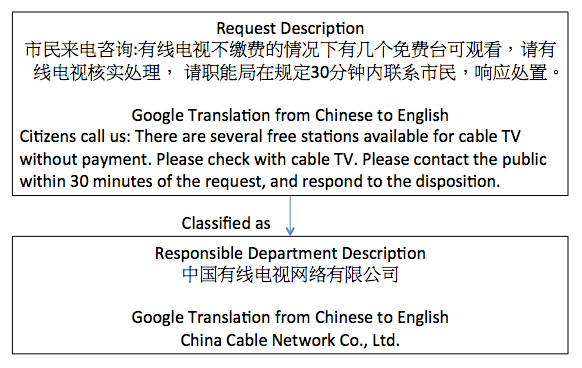}
	\caption{One data sample with features of request description and responsible department description}
	\label{fig:datasample}
\end{figure}

The feature that contains the labeling information is \textit{responsible department description}. A simplified illustrating sample is depicted in Figure~\ref{fig:datasample}. The description needs further processing to generate labels for training and inference. The reason is the description usually contains location phrases with various levels of metropolitan granularity, national, provincial, county, and local communities. This means the records with the same responsible department but different location nouns become separate classes. As a result, the dataset distribution over the labels is spread widely with a large number of classes and the density of classes is diluted. Such a circumstance degrades the training quality and inference accuracy.
 
\vspace{-0.3in}

\begin{figure}[h]
	\centering
	\includegraphics[width=0.6\textwidth]{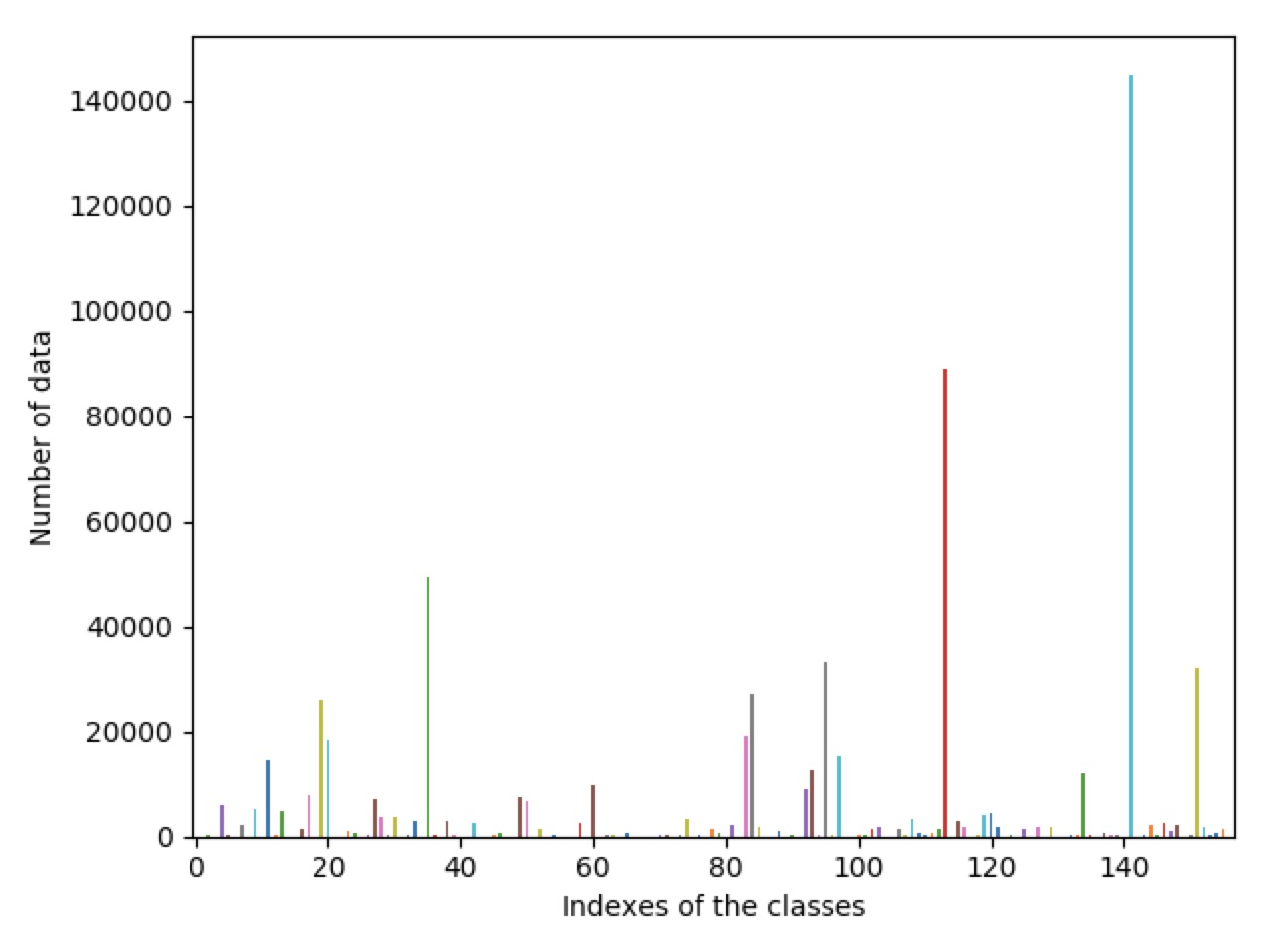}
	\caption{Data distribution statistic for 157 classes, the x-axis is the index of the classes and the y-axis is the total number of data of a class}
	\label{fig:dataDistribution}
\end{figure}

To solve the above problem, we separate the location nouns from department names and titles. Only the organization names and titles remain to generate training labels. For cases that the department names and tiles are in abbreviation, we set up a dictionary and manually create an entry mapping between the standard full name and any forms of variation including abbreviation. This leads to 157 unique department names and titles, which are considered as the labels for classification. We further plot the data sample distributions over the 157 classes in Figure~\ref{fig:dataDistribution}. Among them, there are 101 classes whose data sample sizes are below 1000 samples (approximately 1000 out of a portion of 0.15\%). 

We further explore the dataset characteristics upon observations of data distribution. We develop an inverted index of words in each request description record. The Bag-of-Words (BoW) algorithm \cite{harris1954distributional} is used to build the word pairs with their word counts per record. The word and the sample become a vector that is stored. Such a vector allows us to trace the word occurrence in samples. Therefore we compute the statistics of sample counts grouped by the word occurrence frequency.

\begin{figure*}
	\centering
	\subfigure[Data sample distribution over word frequency in request description]{
		\includegraphics[width=0.35\textwidth]{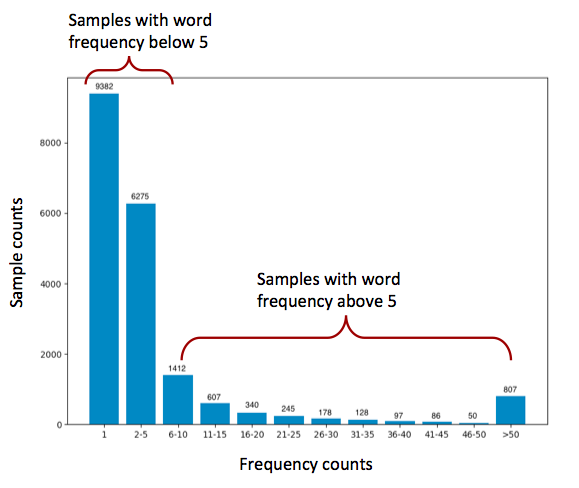}%
		\label{fig:wordfrequencydistribution}
	}\hspace{0.2cm}
	\subfigure[Label distribution over word frequency]{%
		\includegraphics[width=0.4\textwidth]{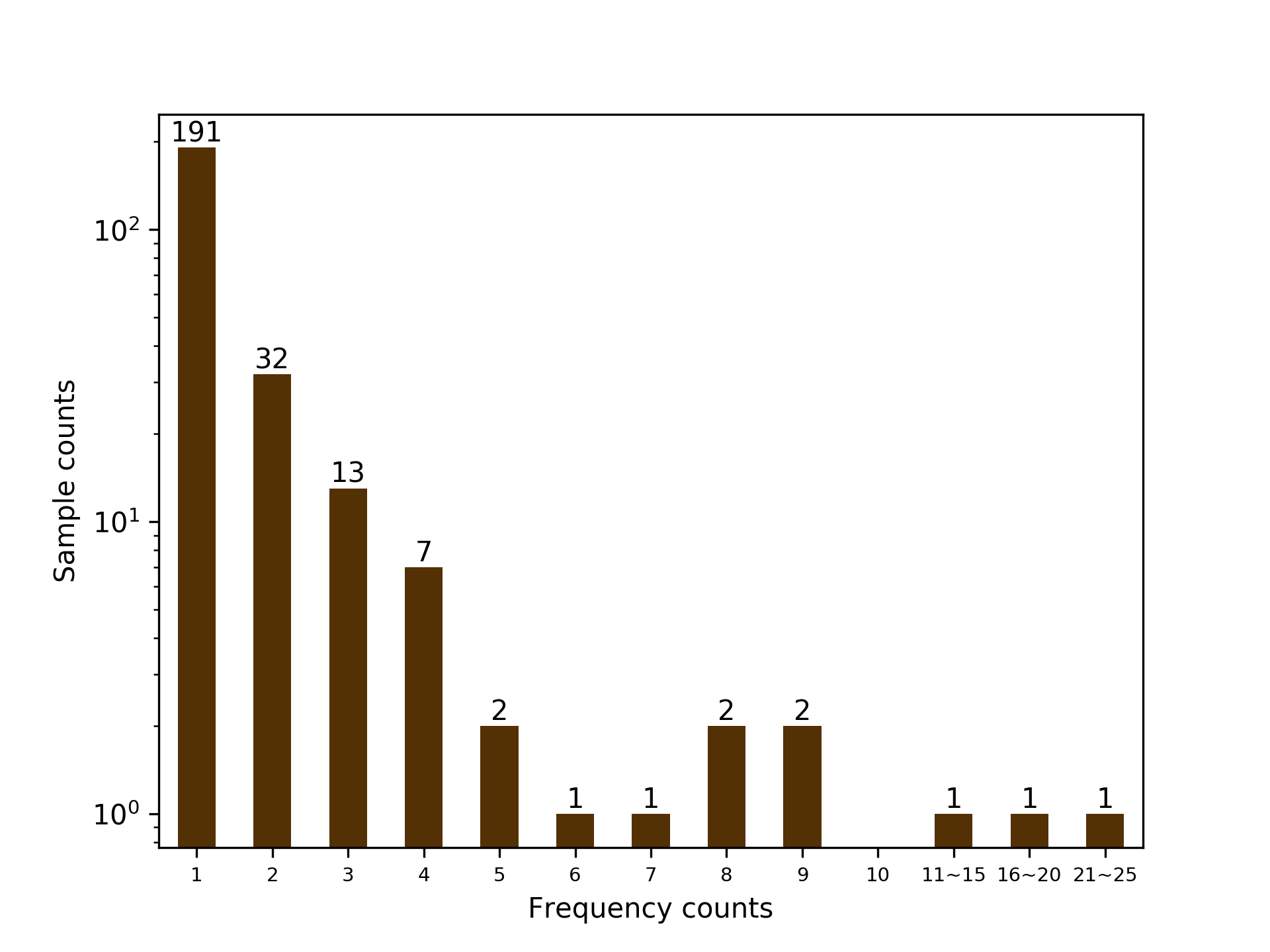}
		\label{fig:labelfrequencydistribution}
	}  
\caption{Data and label distribution}
\end{figure*}

Figure~\ref{fig:wordfrequencydistribution} plots the word frequency distribution. This plot is interrupted as 9382 data samples contain words that only occur once in the whole request description text; 6275 data samples contain words occur 2 to 5 times in the whole request description text. The words of high frequency (such as over 50) only appear in a limited number of samples (such as 807 samples). Since stopping words are already filtered, this plot in Figure~\ref{fig:wordfrequencydistribution} indicates words with low frequency should be further measured with their relevance of other words in the feature space. Likewise, we produce the plot of word frequency and distribution of 157 labels as shown in Figure~\ref{fig:labelfrequencydistribution}. Our solution is training the Word2Vec model to generate the word embedding that represents the relations of word tokens in a high dimensional space. The details are presented in Section~\ref{subsec:featureextraction}.

\subsection{Information Values of Features} 
\begin{figure}[h]
	\centering
	\includegraphics[width=0.7\textwidth]{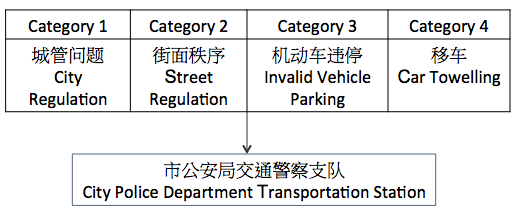}
	\caption{Data sample of categories}
	\label{fig:categorysample}      
\end{figure}

Information values and Weight-of-Evidence (WoE) are techniques of feature selection. Information values measure the prediction power of a feature. The decision to select the four categories of responsible departments as features is evaluated by their information values. The value of each category are tags edited by customer service operators. A data sample is depicted in Figure~\ref{fig:categorysample}. The bottom textbox shows the corresponding responsible department. The four categories represent four levels of tags as a whole capturing context of the request. The tokenization and segmentation workflow is applied to each category.  We combine tags of four categories as a combo for the information value analysis. By statistics, there are 418 unique values of category tag combo. 

\[ [tag_{1}, tag_{2}, ... tag_{n} ], n = 418 \]  

The information values (IVs) are calculated to measure the prediction power of the category tag combo to the class label as the responsible department. Therefore, the 157 responsible departments and the 418 category tag combo form a $157 \times 418$ vector.  Each entry of this vector is notated as $TagCombo_{i,j}$, which represents the counts of data samples with tag combo $j$ that belong to responsible department $i$. Then the notation of $Non-TagCombo_{i,j}$ represents the total counts of data samples with tag combo $l\ne j$, that is

\[ Non-TagCombo_{i,j} =  \sum_{l=1,l \ne j}^{418} \#tag-combo_{i,l}     \]

\begin{figure}[h]
	\centering
	\includegraphics[width=0.7\textwidth]{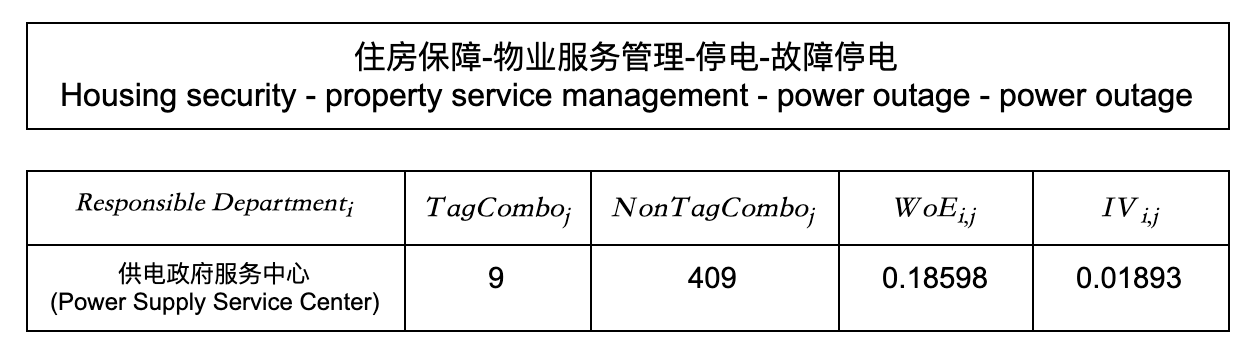}
	\caption{Sample Calculation of IV}
	\label{fig:ivcalculation}      
\end{figure}

Now, we can calculate the value of WoE as 

\[WOE_{i,j} = ln(\frac{TagCombo_{i,j}}{Non-TagCombo_{i,j}})\]
Hence, we calculate the information value vectors for each category tag combo $j$ for responsible department $i$ as shown in Figure~\ref{fig:ivcalculation}.

Finally, the IV values of each category tag combo are summed up for all 157 responsible departments to measure the prediction power of each tag combo as  

\[ IV_{i,j} = (TagCombo_{j}\% - Non-TagCombo_{j}\%) * WOE_{i,j}\] 
\[ IV_{TagCombo_{j}} =  \sum_{i=1}^{157} IV_{i,j}     \]

%

According to the rule of thumb described above, we find only 13 out of 418 category-combos are weak predictions, that is IV value is in the range of (0.02,0.10]. The rest of category-combos have IV values below 0.02, which indicates not useful for prediction. In another word, it means category-combo is not an important feature for classifying responsible departments. They are not selected as input features for classifiers. 

\subsection{Feature Extraction} \label{subsec:featureextraction}

Feature extraction is to produce the word tokens into word vectors with numerical values. In this paper, we apply two methods namely Term-frequency-inverse document frequency (TF-IDF) \cite{sparck1972statistical}, and Word2Vec \cite{mikolov2013efficient}.

\subsubsection{Generating word vector using TF-IDF}
TF-IDF measures the relevance of words, not frequency. That is, word counts in the inverted index discussed in Section~\ref{sub:datadistribution} are replaced with TF-IDF relevance weight across the whole dataset. With TF-IDF, the more samples a word appears in, the less valuable that word is as a signal to differentiate a given request. The relevance weight of a word is calculated in Eq~\ref{eq:weight}.

\begin{equation}\label{eq:weight}
w_{i,j} = tf_{i,j} \times \log (\frac{N}{df_{i}})
\end{equation}

where $N$ is the number of samples;  $tf_{i,j}$ represents the number of occurrence of word token $i$ in sample request $j$; $df_{i}$ represents the number of occurrence of request samples that contain the token $i$. The vector of $w_{i,j}$ is a normalized data format that adds up to one for all the samples. This TF-IDF vector is used as input to the Naive Bayes classifier discussed in Section~\ref{sub:bayesian}. 

\subsubsection{Word embedding using Word2Vec}\label{word2vec}
Word embedding maps word tokens of varied length to a fixed-length word vector as inputs to machine learning models. In nature, word embedding reconstructs linguistic contexts of words and produces a vector space. The algorithm of Word2Vec uses a group of related models that are two-layer neural networks that are trained over a large corpus of text and produces a vector space, typically of several hundred dimensions. Each unique word in the corpus is assigned a corresponding vector in the vector space so that words that share common contexts in the corpus are located in close proximity to one another in the space \cite{mikolov2013efficient}.

We have trained the Continuous Bag-of-Words structure (CBOW) of the Word2Vec model with a corpus collected from the whole dataset. We segment the whole dataset of 65,9421 samples with sentences into data blobs of every five continuous words as input. The central word is the target for output. Empirical experiments indicate using the training dataset as corpus produces a better classification performance. The details of the experiments are not central to the research scope of this paper, thus omitted. 

By statistics, we observe the word length in a request description of the dataset ranges from 1 to 780 with a weighted average of 46.  Over 90\% of the request descriptions have less than 100-word tokens. Thus we set the word vector dimension as 100, and pad zero if the word length is less than 100.

\section{Hierarchical Classification Method}\label{sec:Hierarchical}
A hierarchical classification method handles classification of a large number of possible classes \cite{Silla2011}.  The current training dataset contains 157 unique labels and thus 157 classes. We develop a hierarchical classification method to evaluate whether it works for our dataset. There are two kinds of hierarchical classification. One uses meta-classes as a two-hierarchy structure where leaf classes are grouped by similarity into intermediate classes (the meta-classes)~\cite{hao2007hierarchically}. The other one copes with a pre-defined class hierarchy, a type of supervised learning. In this paper, our method is the former case by means of which we build the hierarchy during the training by a clustering method, and then classify a sample from the meta-classes to the leaf-classes. 

\subsection{Meta-class Generation using K-means and GMM}
To create meta-classes for 157 labels according to similarity, we first apply the K-Means clustering algorithm~\cite{kanungo2002efficient}. The K-Means algorithm first assigns each sample to the cluster whose mean values have the least squared Euclidean distance. Secondly, it calculates the new means to be the centroids of the observations in the new clusters. Finally, the algorithm converges when the assignments no longer change. K-Means applies hard clustering by assigning data points to a cluster. This implies a data sample is assigned to the closest cluster. K-Means is simple to train but does not guarantee convergence to the global optimal whereby degrades the clustering accuracy.

The Gaussian Mixture Model (GMM)~\cite{bilmes1998gentle} is a finite mixture probability distribution model. The parameters of GMM are estimated iteratively by using the Expectation-Maximization (EM) algorithm~\cite{bilmes1998gentle}. GMM/EM determines a sample's probability of belonging to a cluster, which is a soft clustering process.  Each cluster, therefore, can have different options to constrain the covariance such as spherical, diagonal, tied or full covariance, instead of only spherical in K-Means, which means the clustering assignment is more flexible in GMM than in K-Means. The EM algorithm has its limitation. One issue is the number of mixtures affects the overall performance of EM and this number is an unknown prior. Therefore, the optimal number of mixtures is important to ensure an efficient and accurate estimation. 

\begin{figure}[h]
	\centering
	\includegraphics[scale=0.2]{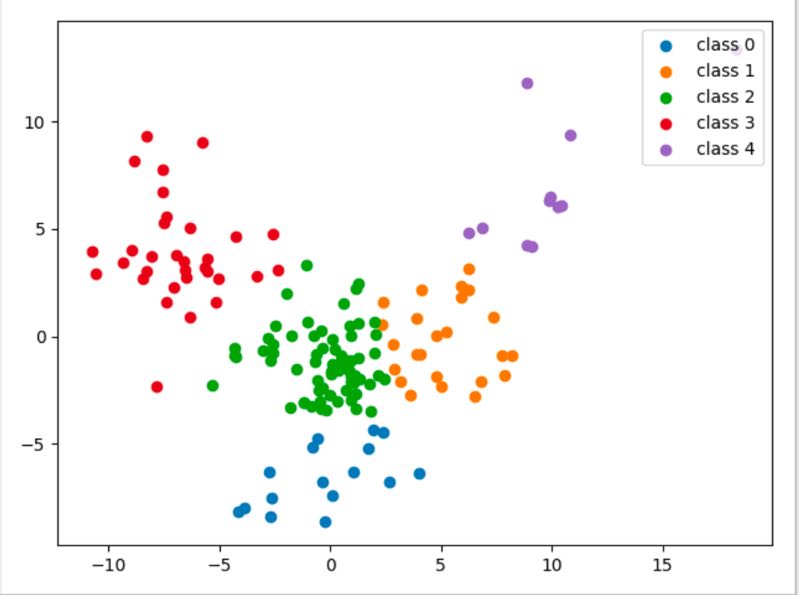}
	\caption{The plot of K-means and GMM clustering}
	\label{fig:kmeans}
\end{figure}

Our solution is applying K-Means to obtain values of centroids (or geometric centers), then initializing GMM with centroids values\cite{figueiredo2002unsupervised}. The input to K-means is a word vector with 100 dimensions of 157 class labels produced from the feature extraction process described in Section~\ref{sub:featureextraction}. We apply silhouette coefficients to select the optimal value of $K$. A silhouette coefficient of a range of [-1, 1] indicates (1) a sample is far away from the neighboring clusters with the value near +1; (2)or a sample is on or very close to the decision boundary between two neighboring clusters with the value of 0; (3) or a sample might have been assigned to the wrong cluster if the value is negative. Figure~\ref{fig:kmeans} plots the clustering result with the optimal $K$ value as 5. We derive the silhouette coefficients by setting the $K$ value range in $[3,10]$. After the silhouette analysis, we consider $K=5$ is the optimal value for K-Means/GMM/EM clustering. Figure~\ref{fig:kmeans} plots the 5 clusters of 157 labels.  

\subsection{Meta-class Generation using Topic-based Clustering}
\begin{figure}[h]
	\centering
	\includegraphics[scale=0.4]{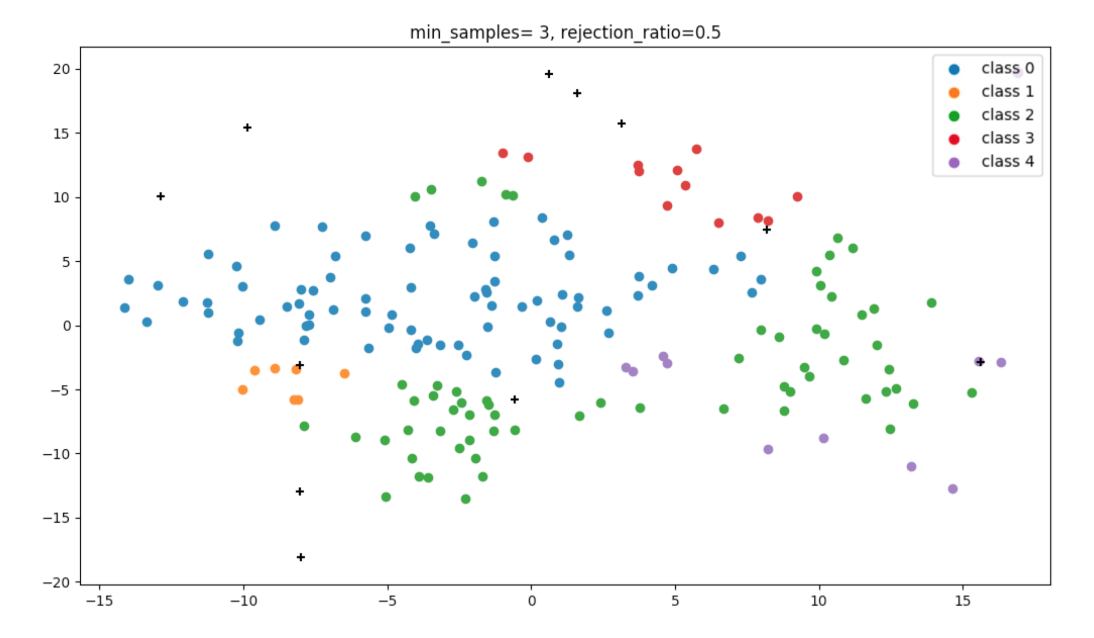}
	\caption{Topic based clustering result}
	\label{fig:lda}
\end{figure}

We develop another clustering method that considers the  157 labels as topics to cluster them into groups with similar themes. In this method, we apply the clustering method of OPTICS (Ordering Points To Identify the Clustering Structure Closely) \cite{ankerst1999optics} that finds a core sample of high density and expands clusters from them. The output from OPTICS provides $K$ clusters. In our method, we apply LDA (Latent Dirichlet Allocation) to output $K$ topics. Each topic consists of a fixed size of words and their associated weights. LDA assumes the Dirichlet prior distribution of topics. In practice, the Dirichlet prior distribution assumes documents (or labels in our case) cover only a small set of topics and topics consist of a small set of words frequently. Finally, we assign labels to a cluster by an entropy-based algorithm.  The algorithm is listed below. The input to this topic-based algorithm is the output of the word vector representation of the 157 labels in  $157 \times 10$ dimensions. The output is $K$ clusters of 157 labels. The clustering result is shown in Figure~\ref{fig:lda}.
\begin{algorithm}
	\DontPrintSemicolon
\SetAlgoLined
\KwInput{$\hat{k} \gets \{k_{j}\}, j \in [1,157]$
	word vector of 157 labels} 

\KwOutput{Clusters of 157 labels}

\begin{algorithmic}
	\STATE $\varsigma \gets$ number of clusters output from OPTICS on Input\; 
	\STATE $num\_topics \gets \varsigma$ to initialize LDA;
	\STATE $ \hat{v_{t}} \gets $ output from LDA; \textcolor{blue} {\COMMENT{a topic vector of $\varsigma \times 10$; each entry is a tuple $t$ of [word: weight]} }
	
	\ForEach{$\hat{v_{t}}[i], 	i 	\in [1,\varsigma]$}{
		
		\STATE $l \gets |\hat{v_{t}}[i]|$; \textcolor{blue} {\COMMENT{number of tuples} }
		
		\STATE ${c_{i}}= \frac{1}{\left | l \right |} \sum_{t \in \hat{v_{t}}[i]} t.weight$ \textcolor{blue}{\COMMENT{calculate the centriod}}
		
		\ForEach{$k_{j}, j \in [1,157]$}{
			\STATE $ p(k_{j}) =  \frac{sim(k_{j},{c}_{i})}{\sum_{r=1}^{ \varsigma} sim(k_{j},{c}_{r})}$
			
			\STATE $e^{c_i}_{k_{j}} = - p(k_{j}) \cdot log(p(k_{j}))$ \textcolor{blue}{\COMMENT{calculate the entropy of $k_{j}$ to a topic $\hat{v_{t}}[i]$}}	 		
		}
		
	}
	\STATE Assign $k_{j}$ to the topic cluster $m$ whose entropy  $e^{c_m}_{k_{j}}$ is minimal;
	
\end{algorithmic}

\caption{Topics and Entropy Based Clustering}
\end{algorithm}

\vspace{-0.3in}
\subsection{Hierarchical Classification}
\begin{figure}[h]
	\centering
	\includegraphics[scale=0.5]{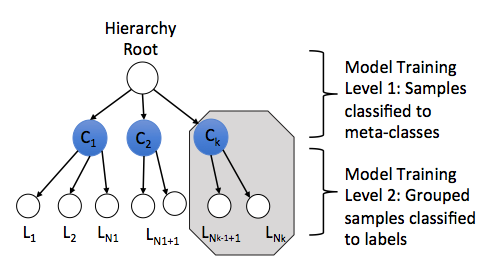}
	\caption{The two-level hierarchy of classes}
	\label{fig:hierarchytree}
\end{figure}

The clustering algorithms generate the meta-classes of 157 labels for training a classifier. The classification is of the structure of a two-level tree as depicted in Figure~\ref{fig:hierarchytree}. The leaves are 157 labels and the non-leaf nodes are the $K$ meta-classes.

A classifier is first trained with the meta-classes, noted as $Model_{meta}$. The classification decides the meta-class that a sample belongs to. As a result, the original data samples are grouped into $K$ clusters. Furthermore, each meta-class $i$ contains $L_{i}$ labels and $\sum_{i=1}^{K} L_{i} = 157$. The data samples classified to one specific meta-class are used again to train a sub-model to further classify these samples to the leaf classes. The hierarchical training produces one meta-class model, noted as $Model_{meta}$ and $K$ number of leaf-class models, one for each cluster. In total, we have $K+1$ models.

When it comes to inference, there are two methods. The first method inputs the test sample to  $Model_{meta}$. Based on the classification on the meta-class, the test sample is further fed to one of the leaf-class models. The second method directly inputs the test samples to each of the leaf-class models. The classification with the highest probability is selected to be the classification result. The first method has shorter inference time as it runs two models, while the second methods run $K$ models. In term of accuracy, the second method tends to be more accurate as it reduces the error propagation from the meta-class level. 

\subsubsection{Hierarchical Naive Bayes Classifier} \label{sub:bayesian}
Both the meta-class model and leaf-class model use the Naive Bayes classifier. The difference is the meta-class model has five classes for classification as the results of topic clustering of the labels. 

In the context of the inputs, $X$ contains the word tokens of the request description. Further, we adopt the Bernoulli Naive Bayes classifier to encode the input of word tokens. The Bernoulli Naive Bayes classifier assumes each feature has only binary values. In this case, the word token is encoded as a one-hot bag of words model that means $1$ indicating the presence of the term or $0$ indicating absence of the term. This leads to 19,607 dimensions of input features from the request description of 40,000 training set. The limitation of this approach is that when the training set changes, the input needs to be encoded again. For example, 80,000 training set leads to 32,698 dimensions of word token one-hot encoding. To avoid the zero value of $P(x_{i}|y_{j})$ causing the posterior probability always being zero, 0.2 smooth value is added to each conditional probability $P(x_{i}|y_{j})$. $y$ represents a set of 157 responsible departments \{$y_{1}, ...y_{157}$\}. The classification is calculated as the class $y_{j}$ that produces the maximum probability given the feature set of X.

\begin{equation}\label{eq:bayesinappro}
\begin{aligned}
P(\widehat y|x_{0},...x_{n}) \propto \max_{j=1}^{157} P(y_{j}) \cdot \prod_{i=1}^{n}P(x_i|y_{j}) \\
\widehat{y} = \arg\max_{j=1}^{157} \prod_{i=1}^{n}P(x_i |y_{j}) \cdot P(y_{j})
\end{aligned}
\end{equation}

\subsubsection{Hierarchical MLP Neural Network} 

By applying a neural network model to the task of text classification, we assume the complex function created using the network of neurons is close to the true relationship between the input features (such as word tokens of request description) and the output (in our case the 157 classes of responsible departments). In other word, a neural network with a certain number of hidden layers should be able to approximate any function that exists between the input and the output according to the  Universal Approximation Theorem~\cite{Cybenko1989}. 

We develop a Multiple Layer Perceptron neural network classifier in the hierarchical model. In this model, both the meta-class model and the leaf-class model have the same network structure as listed in Table~\ref{tab:fullyconnectednetwork} except that the output layer of the meta-class is 5 instead of 157. Accordingly, the weight size is $128 \times 5$ of the meta-class model. Within this structure, each input neuron is connected to each output neuron in the next layer, referred to as a Dense layer. Given the input to the first Dense layer is of the size of $10,000 = 100 \times 100$, the output of the Dense layer is $512$, then the size of weights of this Dense layer is $10,000 \times 512$. We stack two Dense layers with one output layer of 157 classes. The network structure is listed in Table~\ref{tab:fullyconnectednetwork}.

\vspace{-0.3in}
\begin{table}[h]
	\centering
	\caption{The Structure of Fully Connected Neural Network}
	\begin{tabular}{l|c|c|c}
		\hline
		Layers & Input Size& Output Size & Kernel (Weight) Size \\
		\hline  
		\multirow{2}{*}{Dense} &\multirow{2}{*}{10,000}&\multirow{2}{*}{512} &\multirow{2}{*}{$
			10,000\times512
			$  }\\ &&&\\
		\hline
		\multirow{2}{*}{Dense} &\multirow{2}{*}{512}& \multirow{2}{*}{128}&\multirow{2}{*}{$
			512\times128
			$  }\\ &&&\\
		\hline
		\multirow{2}{*}{Output} &\multirow{2}{*}{128}& \multirow{2}{*}{5}&\multirow{2}{*}{$
			128\times5
			$  }\\ &&&\\
		\hline
	\end{tabular}
	\label{tab:fullyconnectednetwork}       
\end{table}

\vspace{-0.4in}
\section{Hybrid Machine Learning}\label{sec:hybrid}
\vspace{-0.1in}

  This hierarchical classification method is useful to deal with a large number of classes by means of training a classifier for a much smaller number of classes (at the level of meta-classes) while keeping comparable levels of confidence. The main issue with hierarchy classification is error propagation. Since the error in the meta-class level classification propagates to the leaf-class classification directly.

  \begin{figure}[h]
  	\centering
  	\includegraphics[scale=0.25]{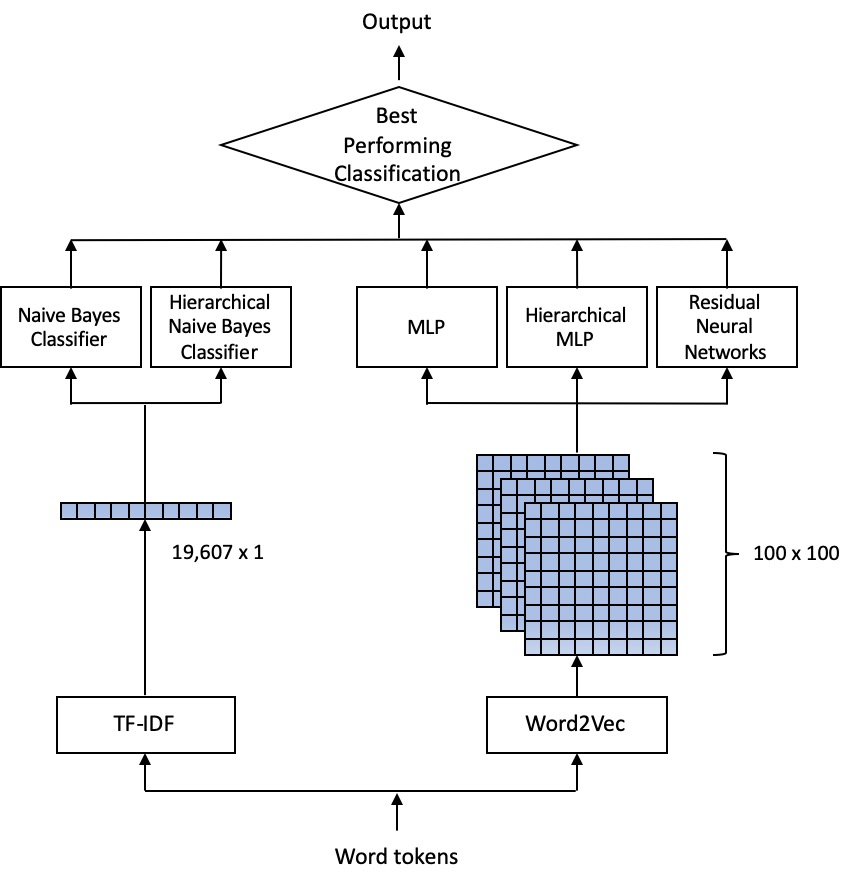}
  	\caption{Oveview of hybrid machine learning models with word embeddings}
  	\label{fig:hybrid}
  \end{figure}
  
  To improve the classification performance, we further develop a residual neural network classifier inspired by ResNet~\cite{ResNet}. By ensembling, the hierarchical model and the residual neural network model, the structure of a hybrid learning method is depicted in Figure~\ref{fig:hybrid}. We add in two basic models as benchmarks Naive Bayes classifier, and an MLP classifier. In totally, we have 5 classifier outputs on the same datasets. A simple ensembling method selects the best performing classification results according to metrics of \textit{log loss}. Classifiers based on Naive Bayes use inputs from word embedding of TF-IDF as a 19,607-dimension word vector. Other three neural network models based on Word2Vector embedding as $100*100$ dimension vectors. 
  
\subsection{Residual Convolutional Neural Network}
\begin{figure}[h]
	\centering
	\includegraphics[scale=0.4, angle=90]{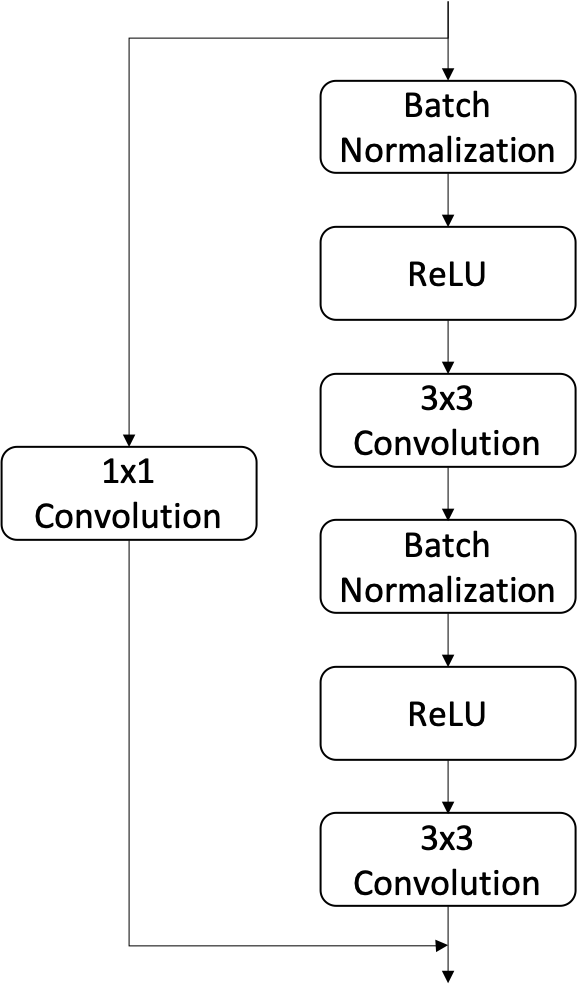}
	\caption{Structure of the Residual Block}
	\label{fig:residualblock}
\end{figure}

In this paper, we apply the \textit{Full pre-activation} structure proposed by He et al.~\cite{he2016identity} to build our convolutional layers. To minimize the gradient vanishing problem when a CNN model grows with deep layers,  residual skip connections or identity mapping are added to convolutional layers. The input to a layer $F(X)$ is added to convolutional output as a combined input to the next layer, $y=F(X, \{W_{i}\}) + W_{s}X$. This structure allows the gradient to be propagated without loss of representations. It is considered that the skip connection and the convolutional layer together form a Residual Block layer. In our model, a Residual Block contains two convolutional layers as shown in Figure~\ref{fig:residualblock}. 

A \textit{Batch Normalization-Activation} layer is placed in-between two convolutional layers.  As discussed in the paper~\cite{he2016identity}, $1 \times 1$ convolution can be useful when there are fewer layers, thus we choose the $1 \times 1$ convolution layer as the skip connection method. Based on the Residual Block structure, we build our 20-layer residual convolutional neural network model shown in Table~\ref{tab:architecture}.


\begin{table}[h]
	\centering
	\caption{Structure of Residual CNN}
	\begin{tabular}{l|c|c}
		\hline
		Layers& Kernel  & Output Size \\
		\hline
		
		\multirow{2}{*}{Convolution} &\multirow{2}{*}{$
			\left[
			\begin{matrix}	
			\ 3\times 3, 32 \ 
			\end{matrix}
			\right] \times 1
			$  }& \multirow{2}{*}{100$\times$100 }\\ & &   \\
		\hline
		\multirow{4}{*}{Residual Block} &\multirow{4}{*}{$
			\left[
			\begin{matrix}
			\ 3\times 3, 32 \ \\ 
			\ 3\times 3, 32 \
			\end{matrix}
			\right] \times 1
			$  }& \multirow{4}{*}{100$\times$100 }\\ & &  \\ & & \\ & & \\
		\hline
		\multirow{4}{*}{Residual Block} &\multirow{4}{*}{$
			\left[
			\begin{matrix}
			\ 3\times 3, 64 \ \\ 
			\ 3\times 3, 64 \
			\end{matrix}
			\right] \times 2
			$  }& \multirow{4}{*}{50$\times$50 }\\ & &  \\ & & \\ & & \\
		\hline
		\multirow{4}{*}{Residual Block} &\multirow{4}{*}{$
			\left[
			\begin{matrix}
			\ 3\times 3, 128 \ \\ 
			\ 3\times 3, 128 \
			\end{matrix}
			\right] \times 2
			$  }& \multirow{4}{*}{25$\times$25 }\\ & &  \\ & & \\ & & \\
		\hline
		\multirow{4}{*}{Residual Block} &\multirow{4}{*}{$
			\left[
			\begin{matrix}
			\ 3\times 3, 256 \ \\ 
			\ 3\times 3, 256 \
			\end{matrix}
			\right] \times 2
			$  }& \multirow{4}{*}{13$\times$13 }\\ & &  \\ & & \\ & & \\
		\hline
		\multirow{4}{*}{Residual Block}&\multirow{4}{*}{$
			\left[
			\begin{matrix}
			\ 3\times 3, 512 \ \\ 
			\ 3\times 3, 512 \
			\end{matrix}
			\right] \times 2
			$  } & \multirow{4}{*}{7$\times$7 }\\ & &  \\ & & \\ & & \\
		\hline
		\multirow{2}{*}{Dense}&\multirow{2}{*}{ } & \multirow{2}{*}{157 }\\ & &   \\
		\hline
		
	\end{tabular}
	\label{tab:architecture}       
\end{table}

\vspace{-0.1in}
\section{The Evaluation}\label{sec:evaluation}

The evaluation focuses on the assessment of the classification performance. We compare the metrics of training five models with different sizes of training data and testing the models with data collected at different time spans. 

\subsection{The Data Setup}
The dataset contains 659,421 samples. We split the data with a ratio of 80\%:20\%  for the training set and the test set. We further partition the training set into shards of 40,000 samples for each shard. Likewise, we partition the test set into shards with 4,000 samples in each shard. Hence, the usage of the dataset is in the unit of a shard. We set up experiments of running 5 models on two data settings: (1) one shard of training set with one shard of test set; and (2) two shards of the training set with one shard of the test set. 

In addition to the 659,421 samples, we also test the best performing model using the data collected in a different time period that contains 35,663 valid samples.  

%
%
%
%

\subsection{The Model Assessment}
The evaluation defines model assessment metrics as Precision and Recall. For a multi-classes classification model, the Precision and Recall score should be calculated for each class and take the average score as the final score. There are two averaging methods, $micro$, and $macro$. The macro method considers every class has the same weight, while in the micro method every data has the same weight. The calculation is as shown below, $P_{i}$ is the Precision of the $i$th class, and $R_{i}$ is the Recall of the $i$th class. $l$ is the total number of classes. 

\begin{equation}
\begin{aligned}
P_{macro}&= \frac{1}{l}\sum_{i=1}^{l}P_i \quad P_i= \frac{TP_i}{TP_i+FP_i}\\
R_{macro}&= \frac{1}{l}\sum_{i=1}^{l}R_i \quad R_i= \frac{TP_i}{TP_i+FN_i}\\
P_{micro}&= \frac{\sum_{i=1}^{l}TP_i}{\sum_{i=1}^{l}TP_i+\sum_{i=1}^{l}FP_i}\\
R_{micro}&= \frac{\sum_{i=1}^{l}TP_i}{\sum_{i=1}^{l}TP_i+\sum_{i=1}^{l}FN_i}
\end{aligned}
\end{equation}

where

$True Positive (TP)$: the real label is positive and the predicted label is also positive;

$False Positive (FP)$: the real label is negative and the predicted label is positive;

$True Negative (TN)$: the real label is negative and the predicted label is also negative;

$False Negative (FN)$: the real label is positive and the predicted label is  negative.\\

The above metrics focus on if the classification is correct or not. Log Loss measures the distance between the predicted label and the real label. It takes the prediction probability for each class of a model as the input and outputs a log loss value as calculated in Eq~\ref{eq:logloss}. The lower the log loss value (such as close to zero), the better performance of a model is. $l$ is the total class number, $y_i$ is the real label and $p_i$ is the predicted probability of class $i$.

\begin{equation}\label{eq:logloss}
Log Loss = \sum_{i=1}^{l}y_ilog(p_i) + (1-y_i)log(1-p_i)
\end{equation}

\subsection{The Experiment Results}
\vspace{-0.1in}
The first set of experiments evaluate the classification performance of the five models. The training data sets are of one shard (40,000 samples) and two shards (80,000 samples) respectively. The test data set is one shard of 4,000 samples. Both the training set and test set are from data samples collected within the same span of time. Table~\ref{tab:157clses} list the metrics measured for 5 classification models. 

\vspace{-0.2in}
\begin{table}[h]
	\centering
	\captionof{table}{Classification Performance on Five Models}
	\begin{tabular}{|l|c|c|c|c|c|}
		\hline
		\multirow{3}{*}{Models} & \multirow{3}{*}{Training Subset} & \multicolumn{4}{|c|}{Metrics} \\
		\cline{3-6} && \multicolumn{2}{|c|}{Precision} & \multicolumn{2}{|c|}{Recall} \\ \cline{3-6}
		&&Micro&Macro&Micro&Macro\\
		\hline
		\multirow{2}{*}{Hierarchical MLP} & 40,000  &0.633 &0.247 & 0.633 & 0.214 \\
		&80,000 &0.686 & 0.332  & 0.686 & 0.288 \\
		\hline
		\multirow{2}{*}{MLP} & 40,000   &0.662 &0.276 & 0.662 & 0.236\\
		&80,000 &0.689 & 0.281  & 0.689 & 0.233\\
		\hline
		\hline
		\multirow{2}{*}{Hierarchical Naive Bayes} & 40,000  &0.746 &0.439 & 0.746 & 0.367\\
		&80,000 &0.719 &0.375 & 0.719 & 0.288\\
		\hline
		\multirow{2}{*}{Naive Bayes} &40,000  &  0.734 & 0.358 & 0.734 & 0.254 \\
		&80,000& 0.700  & 0.296 & 0.700 & 0.194\\
		\hline 
		\hline 
		\multirow{2}{*}{Residual CNN} &40,000 & 0.754&0.420& 0.754& 0.389 \\
		&80,000 & \textbf{0.787} & \textbf{0.510}  & \textbf{0.787} & \textbf{0.444}\\
		\hline  
	\end{tabular}
	\label{tab:157clses}       
\end{table}

\textbf{Tranining sample size.} By doubling the training sample size, we observe three models improve the classification performance, including MLP, Hierarchical MLP, and Residual CNN. 

Both Naive Bayes and Hierarchical Naive Bayes metrics decrease. As presented in section~\ref{sub:bayesian}, we obtain 19,607 dimensions of input features from the request description of 40,000 training set and 32,698 dimensions of word token one-hot encoding from 80,000 training set. The observation from this experiment indicates increasing the feature size impacts the performance of our implementation of Naive Bayes classifier. Our Naive Bayes classifier learns one shard of the training set more linearly separable than the doubled size of the training set. In comparison, the word embedding method applied allows the MLP and the Residual CNN classifiers remain the same feature size of $100 \time 100$ dimensions of word token vectors regardless of the size of training data.

\begin{figure}[h]
	\centering
	\includegraphics[scale=0.4]{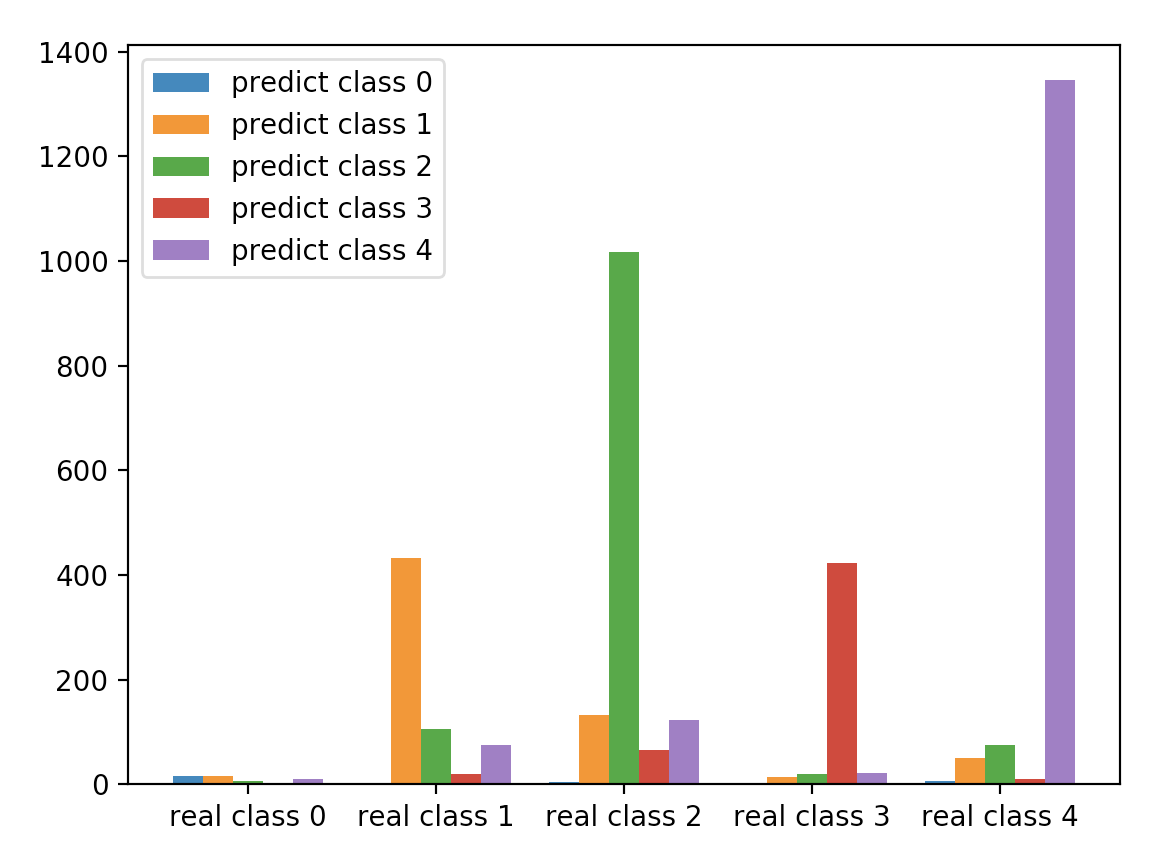}
	\caption{Distribution of predicted classes vs real classes}
	\label{fig:confusion}
\end{figure}

\textbf{Hirarchical vs Non-hierarchical.} Hierarchical classification improves the marginal performance of Naive Bayes classifier. For MLP and Residual CNN, they both perform better than hierarchical classification. In all experiments, Residual CNN outperforms other models. Hierarchical classification overall has lower performance than non-hierarchical classification. The benefit of introducing meta-class through the hierarchical classification method is observing the source of classification errors. Figure~\ref{fig:confusion} shows the classification results with 5 meta-classes. It indicates the classification errors are mainly from the fact that class 2 and 4 are misclassified to class 1; class 1, 3, and 4 are misclassified to class 2. We also observe from the experiments that all the five classifiers produce over 80\% precision for the 5 meta-class classifications. The precision is higher than the classification performance of 157 classes. Due to the space limitation, we skip the values in details.   

%
%

\textbf{Blind Test.} The second set of experiments run on a blind test. We select the best performing model trained from each of the 5 classifiers and further test them using the whole 35,663 data samples collected from a different span of time than the first set of experiments. The result is shown in Table~\ref{tab:blindtest}. Two classifiers, Naive Bayes and Residual CNN produce better classification performance than other three classifiers. Still, Residual CNN performs the best on the blind test data. 

\vspace{-0.4in}
\begin{table}
	\centering
	\captionof{table}{Classification Performance of Blind Test}
	\begin{tabular}{|l|c|c|c|c|}
		\hline
		\multirow{3}{*}{Models} & \multicolumn{4}{|c|}{Metrics} \\
		\cline{2-5}   & \multicolumn{2}{|c|}{Precision} & \multicolumn{2}{|c|}{Recall} \\ \cline{2-5}
		&Micro&Macro&Micro&Macro\\
		\hline
		Hierarchical Fully connected NN
		&0.650  & 0.244   & 0.650  & 0.192  \\
		\hline
		{Fully connected NN} 
		&0.689 & 0.259  & 0.689 & 0.214\\
		\hline
		\hline
		{Hierarchical Naive Bayesian} 
		&  0.678 & 0.251   & 0.678 & 0.201\\
		\hline  
		{Naive Bayesian} 
		&  0.726 & 0.295   & 0.726 & 0.256\\
		\hline  
		\hline
		{Residual Network} 
		&\textbf{0.764}  & \textbf{0.417}   & \textbf{0.764}  & \textbf{0.352}  \\
		\hline  
	\end{tabular}
	\label{tab:blindtest}       
\end{table}

\textbf{Log loss.} We further evaluate the Residual CNN performance on the blind test data using log loss to measure the distance between the predicted label and the real label. The result is listed in Table~\ref{tab:logloss}. When applied to different test data, the Residual CNN has marginal log loss change. 

\begin{table}
	\centering
	\captionof{table}{Log Loss of Hybrid Classifiers}
	\begin{tabular}{|l|c|c|}
		\hline
		Models& Test Data Size  & Log Loss\\
		\hline
		\multirow{2}{*}{Best Performing Residual CNN} 
		& 4,000   & 1.152    \\
		& 35,663  & 1.192    \\
		\hline		
	\end{tabular}
	\label{tab:logloss}       
\end{table}

\textbf{Inference time.} We further measure the inference time taken on the 4,000 test data set. Note that the classifiers of Naive Bayes and Hierarchical Naive Bayes run on a CPU node while other three neural network models run on a GPU node. Therefore the comparison between Navie Baye models and neural network models should not be evaluated against the absolute values. Instead, we observe the hierarchical classification introduce approximately 10 times inference computing delays. The inference time taken by Residual CNN is over 20 times than MLP.

\begin{figure}[h]
	\centering		
	\includegraphics[scale=0.5]{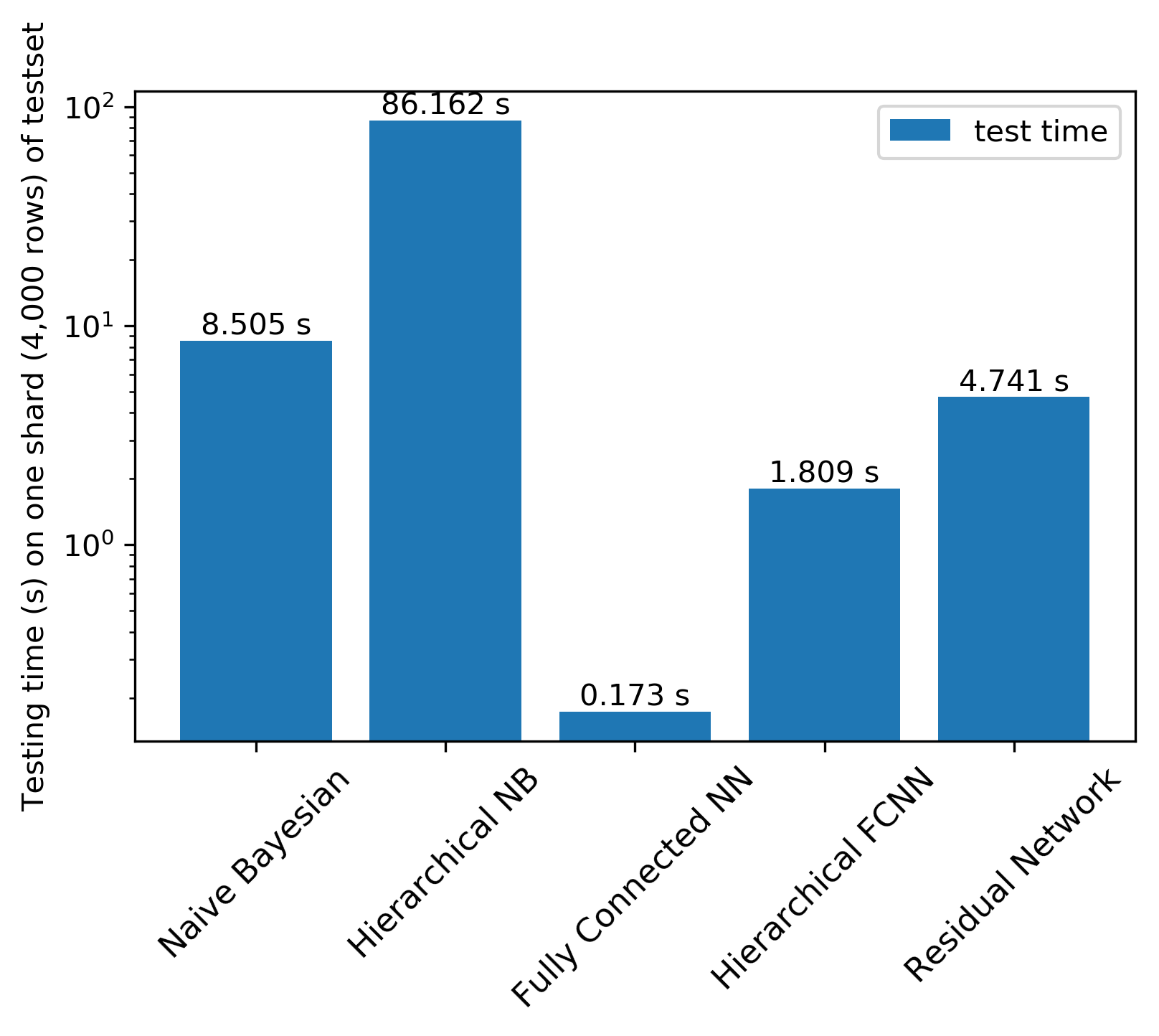}
	\caption{Inference time on the test data of 4,000 samples}
	\label{fig:testingtime}
\end{figure}

\textbf{Summary.} The experiments set up different sizes of training data and test data. The observation shows the residual convolutional neural network model produces the best performance over other classifiers. We also observe that Naive Bayes model with the one-hot encoding of word tokens performs reasonably well with the limitation of handling increasing feature sizes. A simple two-layer fully connected neural network model has the advantage of fast inference time.

\subsection{Threat to Validity}
This paper presents the first stage towards automated smart dispatching of residential service requests. The focus of this paper is exploring a combination of word embedding techniques, and machine learning models to improve classification performance. Our hybrid machine learning model follows a simple ensembling approach for selecting the best performing classifier based on metrics of log loss. For our model to be deployed as an online machine learning service handling requests, the model selection mechanism needs to be in the feedback loop based on actual inference results and quality. A weighted score of multiple metrics that best reflect the online service requirements should be further developed to replace the current simple selection based a single metrics. 

Our evaluation compares with two benchmarking models of Naive Bayes and MLP classifiers. In the literatures, NLP based machine learning methods on news classifications, customer review sentiment analysis, movie review classifications have related work to our method. However, the datasets are specific to the domains without a direct solution to address the problems in our datasets that are not directly labeled for training. Combining our hybrid word embedding and learning models with exiting mining and learning methods become a new stream of investigation that requires a dedicated project to develop that is beyond the current funding budget.

\section{Conclusion}\label{sec:conclusion}
In this paper, we present a machine learning based method of natural language classification task for a real-world application. We carry out a rigorous analysis of the dataset and design a feature engineering process that select and extract features with statistical evidence. We apply two-word embedding techniques and develop five classification models. This hybrid machine learning method produces benefits, namely (1)  generating suitable labels for supervised learning;  (2) clustering data samples into meta-class for training and initializing models to improve classification performance over unbalanced data samples;(3) producing the best performing model through comprehensive experiments and evaluation; and (4) understanding the source of error with the hierarchical classification method.  It remains our future work to explore newly published word embedding model to study the effects of word embedding on classification performance.

\bibliographystyle{spmpsci_unsrt}
\bibliography{reference}   

\end{document}